\documentclass{article}
\usepackage{icml2002}
\usepackage{epsf}
\usepackage{mlapa}
\usepackage{times}
\usepackage{psfig}
\usepackage{epic}
\usepackage{eepic}
\usepackage{amsmath}
\usepackage{multirow}
\usepackage{calc}

\newcommand{\pomdp}{{\sc pomdp}}

\newcommand{\reinf}{{\sc reinforce}}
\newlength{\figwidth}
\newsavebox{\fminibox}
\newlength{\fminilength}

\newcommand{\commentout}[1]{}

\newcommand{\tuple}[1]{\langle #1 \rangle}

\begin{document} 

\twocolumn[
\icmltitle{Learning from Scarce Experience}
\icmlauthor{Leonid Peshkin}{pesha@eecs.harvard.edu}
\icmladdress{Harvard Center for Artificial Intelligence \\
Dworkin Bld. 134, Cambridge, MA 02138}
\icmlauthor{Christian R. Shelton}{cshelton@cs.stanford.edu}
\icmladdress{Stanford Computer Science Department \\
Gates Bld. 133, Stanford, CA 94305}
\vskip 0.3in
]

\begin{abstract}
Searching the space of policies directly for the optimal policy has been one
popular method for solving partially observable reinforcement learning
problems. Typically, with each change of the target policy, its value is
estimated from the results of following that very policy. This requires a
large number of interactions with the environment as different polices are
considered. We present a family of algorithms based on likelihood ratio
estimation that use data gathered when executing one policy (or collection of
policies) to estimate the value of a different policy. The algorithms combine
estimation and optimization stages. The former utilizes experience to build a
non-parametric representation of an optimized function. The latter performs
optimization on this estimate. 
We show positive empirical results and 
provide the sample complexity bound.

\end{abstract}

\section{Introduction}
\label{intro}

Research in reinforcement learning focuses on designing algorithms
for an agent interacting with an environment to adjust its behavior to
optimize a long-term return.  For environments which are fully-observable
({\it i.e.}~the observations the agent makes 
contain all of the necessary information about the state of the environment),
this problem can often be solved using a one-step look ahead analysis
to formulate the solution as a dynamic programming problem.  However,
for the case of partially observable domains ({\it i.e.}~the observations
are stochastic or incomplete representations of the environment's state),
the perceptual aliasing of the observations makes such methods infeasible.

One viable approach is to search directly in a parameterized
space of policies for a local optimum. Following Williams's
{\reinf} algorithm~\cite{Williams92}, searching by
gradient descent has been considered for a variety of policy
classes~\cite{Marbach,Baird99,MeuleauUAI99,SuttonNIPS99,BaxterICML00}. A
commonly recognized shortcoming of all these variations on gradient
descent policy search is that they require a very large number of samples
(instances of agent-environment interaction) to converge.

This inefficiency arises because the value of the policy (or its
derivative) is estimated by sampling from the returns obtained by
following that same policy.  Thus, after one policy is evaluated and a new
one proposed, the samples taken from the old policy must be discarded.
Each new step of the policy search algorithm requires a new set of samples.
The key to solving this inefficiency is to use data gathered when using
one policy to estimate the value of another policy. The method known as
``likelihood ratio'' estimation enables this data reuse.


Stochastic gradient methods and likelihood ratios have been long used for
optimization problems (see work of
\singleemcite{Glynn86,Glynn89,Glynn90,Glynn94}).
Recently, stochastic gradient descent methods, in particular
{\reinf}~\cite{Williams92,Williams91}, have been used in
conjunction with policy classes constrained in various ways:
with external
memory~\cite{Peshkin99}, finite state controllers~\cite{MeuleauUAI99} and in
multi-agent settings~\cite{Peshkin00}.
The idea of using
likelihood ratios in reinforcement learning was suggested by
\emcite{Csaba} and developed for
solving {\sc mdp}s with function approximation by \emcite{PrecupICML00}
and for gradient descent in finite state controllers by
\emcite{MeuleauTR00}.
However, only on-line optimization was considered.
\singleemcite{Shelton01,SheltonPhD}
developed greedy algorithm for combining
samples from multiple policies in normalized estimators and demonstrated
a dramatic improvement in performance.
\emcite{Peshkin01} showed that likelihood-ratio
estimation enables the application of methods from statistical learning
theory to derive {\sc pac} bounds on sample complexity.

\emcite{KearnsNIPS99} provide a method for estimating the return of every
policy simultaneously using data gathered while executing a fixed policy
without the use of likelihood ratios. In some domains, there is a natural
distance between observations and actions which also allows one to re-use
experience without likelihood ratio estimation. \emcite{Glynn00} demonstrate
algorithms for kernel-based {\sc rl} in one such domain: financial planing
and investments.


This paper extends our previous work by presenting a generalized method of
using likelihood ratio estimation in policy search and investigating the
performance of this method under different conditions on illustrative
examples. By this publication we hope to stimulate a dialog between the
communities of reinforcement learning and computational learning theory. We
present a clear outline of all algorithms in a hope to attract wider research
community to applying these algorithms in various domains. We also present
some new bounds on a sample complexity of these algorithms, making an attempt
to relate these results to empirical results.
We begin this paper with a brief definition of reinforcement learning and
sampling in order to clarify our notation. Then we present our algorithm and
consider the question of how to sample. Finally we consider the question of
how much to sample and present a {\sc pac}-style bound as a quantitative
answer.


\section{Background}
\label{notation}

We introduce the environment model and importance sampling
in a single mathematical notation.  In particular, we keep the standard
notation for partially observable Markov decision processes and modify the
sampling notation to be consistent.

\subsection{Environment Model}

The class of problems we consider can be described by the {\em partially
observable Markov decision process} (\pomdp) model. In a \pomdp, a sequence
of events occur for each time step: an agent observes the observation $o(t)
\in {\cal O}$ dependent on the state of environment $s(t) \in {\cal S}$;
it performs an action $a(t) \in {\cal A}$ according to its policy,
inducing a state transition of the environment; then it receives
a reward $r(t)$ based on the action taken and the environment's state.
A {\pomdp} is defined by four probability distributions (and the spaces
over which those distributions are defined): a distribution over starting
states, a distribution over observations conditioned on the state, a
distribution over next states conditioned on the current state and the
agent's action, and a distribution over rewards given the state and action.
These distributions, specifying the dynamics of the environment, are
unknown to the agent along with the state space of process, ${\cal S}$. 

Let $H=\left\{\tuple{o(1),a(1),r(1),\ldots,o(t),a(t),r(t),o(t+1)}\right\}$
denote the set of all possible experiences sequences of length $t$. Generally
speaking, in a \pomdp, a policy $\pi$ is a function specifying the action to
perform at each time step as a function of the whole previous history:
\mbox{$\pi: H \rightarrow {\cal P}(A)$}. This function is parameterized by a
vector $\theta \in \Theta$. {\em Policy class} $\Theta$ is a set of policies
realizable by all parameter settings. We assume that the probability of the
elementary event is bounded away from zero: \mbox{$0 \le \underline{c} <
\Pr(a|h,\theta) < \overline{c} \le 1$}, for any $a\!\in\!A$, $h\!\in\!H$, and
$\theta\!\in\!\Theta$. A history $h$ includes several immediate rewards
$\tuple{r(1), \ldots, r(i), \ldots }$ that are typically summed to form a
return, $R(h)$, but our results are independent of the method used to compute
the return.

Together with the distributions defined by the \pomdp, any policy
\mbox{$\theta \in \Theta$} defines a conditional distribution
$\Pr(h|\theta)$ on the class of all histories $H$. The value of policy
$\theta$ is the expected return according to the probability induced
by this policy on the history space: $ V(\theta)\!=\!{\rm E}_{\theta}
\left[ R(h) \right] = \sum_{h \in H} \left[ R(h) \Pr(h|\theta) \right],
$ where ${\rm E}_{\theta}$ stands for ${\rm E}_{\Pr(h|\theta)}$. We
assume that policy values (and returns) are non-negative and bounded
by $V_{max}$.  The {\em objective} of the agent is to find a policy
$\theta^*$ with optimal value: \mbox{$\theta^* = {\rm argmax}_{\theta}
V(\theta)$}. Because the agent does not have a model of the environment's
dynamics or reward function, it can not calculate $\Pr(h|\theta)$ and
must estimate it via sampling.

\subsection{Sampling}
\label{sampling}

If we wish to estimate the value $V(\theta)$ of the policy $\theta$, we
may draw sample histories from the distribution induced by this
policy by executing the policy multiple times in the environment.
After taking $N$ samples \mbox{$\bar h = \{h_i\}, i \in (1, \dots, N)$}
we can use the unbiased estimator:
$$
\hat V_{\bar h}(\theta) = \frac{1}{N} \sum_i R(h_i)\,\,.
$$
Imagine, however, that we are unable to sample from the policy
$\theta$ directly, but instead have samples from another policy $\theta'$. The
intuition is that if we knew how ``similar'' those two policies were to one
another, we could use samples drawn according to the distribution $\theta'$ and
make an adjustment proportional to the similarity of the policies. Formally we
have: \\ $V(\theta) = \sum_h R(h) \Pr(h|\theta) 
 = \sum_h R(h) 
        \frac{\Pr(h|\theta')}{\Pr(h|\theta')} \Pr(h|\theta)  $ \\ 
\mbox{\hspace{.51cm}   } $
 = \sum_h R(h) \frac{\Pr(h|\theta)}{\Pr(h|\theta')} \Pr(h|\theta') 
 = {\rm E}_{\theta'} \left[ R(h)
        \frac{\Pr(h|\theta)}{\Pr(h|\theta')}\right]\;.
$ \\
Now we can construct an unbiased {\em indirect} estimator for the
distribution $\Pr(h|\theta')$ which is called an {\em importance sampling}
estimator~\cite{Rubinstein81} $\hat V^{\rm IS}_{\theta',\bar
h}(\theta)$ of $V(\theta)$: 
$$
\hat V^{\rm IS}_{\theta',\bar h}(\theta) = \frac{1}{N} \sum_i R(h_i) \frac{\Pr(h_i|\theta)}{\Pr(h_i|\theta')}\,\,.
$$ 
We can normalize the importance sampling estimate to obtain a
lower variance estimate at the cost of adding bias. Such an estimator is
called a {\em weighted importance sampling} estimator and has the form
$$
\hat V^{\rm WIS}_{\theta',\bar h}(\theta) = \frac{1}{\sum_i
\frac{\Pr(h_i|\theta)}{\Pr(h_i|\theta')}} \sum_i R(h_i)
\frac{\Pr(h_i|\theta)}{\Pr(h_i|\theta')}\, ,
$$
which has been found to be better-behaved than $V^{\rm IS}_{\theta',\bar
h}(\theta)$ both theoretically and
empirically~\cite{MeuleauTR00,Shelton01,PrecupICML00}.

Note that both estimators contain the quantity
$\frac{\Pr(h_i|\theta)}{\Pr(h_i|\theta')}$, a ratio of likelihoods. The key
observation for the remainder of this paper is that while an agent is not
assumed to have a model of the environment and therefore is not able to
calculate $\Pr(h|\theta)$, it is able to calculate the likelihood ratio
$\frac{\Pr(h|\theta)}{\Pr(h|{\theta'})}$ for any two policies $\theta$ and
$\theta'$~\cite{SutBar98,MeuleauTR00,PeshkinPhD}. $\Pr(h|\theta)$ can be written as a
product of $\Phi(h)$ and $\Psi(h)$ where $\Phi(h) = \prod_{t=1}^T
\Pr(a(t)|o(1), \dots ,o(t),\theta)$ is the contribution of all of the agent's
actions to the likelihood of the history, and $\Psi(h)$ is the contribution
of environmental events. Because the $\Psi(h)$ component is independent of
the policy ({\it i.e.}~it does not depend on the policy parameter, only on
the history and the {\pomdp} distributions), it cancels from the ratio, and
we have $\frac{\Pr(h|\theta)} {\Pr(h|{\theta'})} =
\frac{\Phi(h|\theta)}{\Phi(h|{\theta'})}$. $\Phi(h|\theta)$ depends only on
the agent, the observations, and the actions (not the states), is known to
the agent, and can be computed and differentiated. This allows us to
construct more efficient learning algorithms that can take advantage of past
experience.

Finally, if the sampling distribution is not constant ({\it i.e.} each sample
is drawn from a different distribution), a single unbiased importance
sampling estimator can be constructed by using all of the samples where the
assumed single sampling distribution is the mixture of the true sampling
distributions. Thus, if samples were taken according to policies $\theta_1,
\theta_2, \dots, \theta_n$, $\Pr(h|{\theta'})$ from above is replaced with
$\frac{1}{N}\sum_j \Pr(h|{\theta_j})$. \emcite{Shelton01} gives more
details for importance sampling estimators with independent, but not
identically drawn, samples. Using this new estimator allows us to change
policies (sampling distributions) during sampling.

\section{Algorithms}
\label{algorithm}

\begin{figure}[h]
\begin{center}
\centerline{\psfig{file=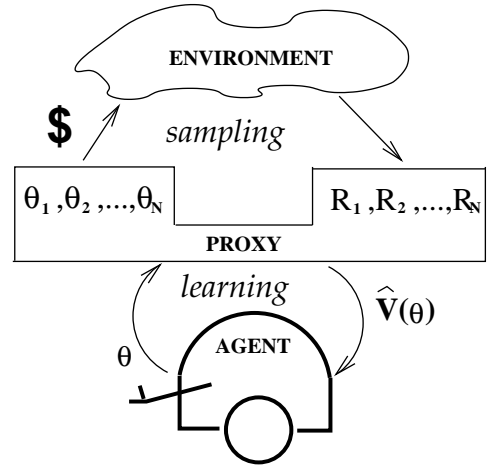,width=6.5cm,angle=0}}
\caption{A diagram of the policy evaluation process.  The sampling
process is costly and therefore only performed a limited number of times.
The proxy collects the samples from the environment and constructs an
agent-centric model that predicts the effects of hypothetical agent
policies.  The agent learns by interacting with the proxy.}
\label{proxy}
\end{center}
\vskip -0.2in
\end{figure}

Consider constructing a {\em proxy environment} that contains a
non-parametric model of the values of all policies as illustrated by
Figure~\ref{proxy}. This model is a result of trying several policies
$\theta_1\ldots\theta_N$. Given an arbitrary new policy $\theta \in \Theta$,
the proxy environment returns an estimate of its value $\hat V(\theta)$ as
if the policy were tried in the real environment. Assuming that obtaining
a sample from environment is costly, we want to construct the proxy module
based only on a small number of queries about policies $\{\theta_i\},
i=1\dots N$ that return values $R_i$. These queries are implemented by the
{\tt sample} routine (Table~\ref{table:sample}). After getting $N$ samples,
it requires memory of size $O(NT(\log|{\cal O}|+\log|{\cal A}|))$ to store
the data, where $T$ is the length of a trial and ${\cal O}$ and ${\cal A}$
are the sets of possible observations and action respectively. However, for
many policy classes this memory requirement can be reduced. For example, if
the policy class is reactive (conditioned on the current observation, the
probability of the current action has no dependence on the past), the history
can be summarized sufficiently by the counts of the number of times each
action was chosen after each observation. This requires memory of size
$O(N\log(T)|{\cal O}||{\cal A}|)$.

\begin{table}[t]
\caption{The {\tt sample} routine accepts a policy parameter setting
and outputs the return and history from one sample of the policy.
Note that in many cases the entire history need not be returned.}
\label{table:sample}
\rule{\linewidth}{1pt}
\begin{center}
\begin{tabbing}
1\=11\=11\=11\kill
 {\bf Input:} policy $\theta$ \\
 {\bf Init:} \\
 \> $h \leftarrow (), R \leftarrow 0$ \\
 \> Get initial observation $o_0$. \\
 {\bf For} each time step $t$ of the trial: \\
 \> Draw next $a_t$ from $\pi(o_t,a_t,\bar \theta)$ \\
 \> Execute $a_t$. \\
 \> Get observation $o_t$, reward $r_t$. \\
 \> $R \leftarrow R + r_t$ \\
 \> $\text{\tt concatenate}(h,(o_t, a_t))$ \\
 {\bf Output:} experience $\tuple{R,h}$
\end{tabbing}
\end{center}
\rule{\linewidth}{1pt}
\vskip -0.2cm
\end{table}

\begin{table}[t]
\caption{The {\tt evaluate} routine computes the proxy's estimate of the
value of a policy and its derivative. The inner loop (over $j$) can be
removed with caching, making the routine faster.}
\label{table:evaluate}
\rule{\linewidth}{1pt}
\begin{center}
\begin{tabbing}
11\=11\=11\=11\kill
 {\bf Input:} policy $\theta$, data $D = \{\tuple{\theta_i,R_i,h_i}\}\
 \mbox{for}\ i \in \{1\dots N\}$ \\
 {\bf Init:} $\hat V \leftarrow 0$, $\Delta \hat V \leftarrow 0$, $\kappa
\leftarrow 0$ \\
 {\bf For} $i = 1$ to $N$: \\
 \> {\bf Init:} $\Psi_i' \leftarrow 0$ \\
 \> {\bf For} $j = 1$ to $N$: \\
 \> \> $\Phi_i' \leftarrow \Phi_i' + \Phi(h_i|\theta_j)$ \\
 \> $\Phi_i \leftarrow \Phi(h_i|\theta)$ \\
 \> $\kappa \leftarrow \kappa + \frac{\Phi_i}{\Phi_i'}$ \\
 {\bf For} $i = 1$ to $N$: \\
 \> $\hat V \leftarrow \hat V + R_i\frac{\Phi_i}{\Phi_i'}$ \\
 $\hat V \leftarrow \frac{\hat V}{\kappa}$ \\
 {\bf For} $i = 1$ to $N$: \\
 \> $\Delta\Phi_i \leftarrow \frac{\partial \Phi(h_i|\theta)}{\partial \theta}$ \\
 \> $\Delta\hat V \leftarrow \Delta\hat V + (R_i-\hat V)\frac{\Delta\Phi_i}{\Phi_i'}$ \\
 $\Delta\hat V \leftarrow \frac{\Delta\hat V}{\kappa}$ \\
 {\bf Output:} proxy evaluation $\hat V$ and derivative $\Delta\hat V$
\end{tabbing}
\end{center}
\rule{\linewidth}{1pt}
\vskip -0.8cm
\end{table}

This proxy can be queried by the learning algorithm as shown in
table~\ref{table:evaluate}.  In response to a policy parameter settings,
the routine {\tt evaluate} returns its estimate of the expected return and
its derivative.  The algorithm shown in table~\ref{table:evaluate} 
computes the weighted importance sampling estimate.
For simplicity, the inner loop (over $j$) is shown.  In practice the
computations in this loop do not need to be redone for every evaluation.
Using memory of size $O(N)$, the values $\Phi'_i$ can be computed ahead of
time (in constant time per sample) thus reducing the evaluation to $O(N)$
time.

The {\tt evaluate} routine relies on two other routines: one to calculate
$\Phi(h|\theta)$ and one to calculate the derivative of $\Phi(h|\theta)$.
Recall that $\Phi(h|\theta)$ is the policy's factor of the probability of the
history $h$. As an example, if we assume the policy to be reactive, the
parameter $\theta_{o,a}$ to be the probability to selecting action $a$ after
observing $o$, and $n_{o,a}$ to be the count of the number of times action
$a$ was chosen after observing $o$ during the history,
\begin{align*}
\Phi(h|\theta) &= \prod_{o,a} \left(\theta_{o,a}\right)^{n_{o,a}} \\
\frac{\partial\Phi(h|\theta)}{\partial\theta_{o,a}} &=
   \frac{n_{o,a}}{\theta_{o,a}}\Phi(h|\theta)\,\,.
\end{align*}
\emcite{MeuleauTR00,PeshkinPhD} describe how to compute these quantities for
reactive policies with Boltzmann distributions and \emcite{Shelton01}
describes how to compute these quantities for finite-state controllers.

\begin{table}[t]
\caption{The {\tt learn} routine accepts the number of trials it is allowed
and returns its guess at the optimal policy. It relies on four external
routines: {\tt pick\_sample} which selects a policy to sample given the data
and the current best guess, {\tt sample} as shown in
table~\protect\ref{table:sample}, {\tt add\_data} which adds the new data
point to the data collected so far, and {\tt optimize} which performs some
form of optimization on the proxy evaluation function.}
\label{table:learn}
\rule{\linewidth}{1pt}
\begin{center}
\begin{tabbing}
11\=11\=11\=11\kill
 {\bf Input:} number of samples/trials $N$ \\
 {\bf Init:} $D \leftarrow ()$, $\theta^* \leftarrow \text{random policy}$ \\
 {\bf For} $i = 1$ to $N$: \\
 \> $\theta \leftarrow \text{\tt pick\_sample}(D,\theta^*)$ \\
 \> $\tuple{R,h} \leftarrow \text{\tt sample}(\theta)$ \\
 \> $D \leftarrow \text{\tt add\_data}(D, \tuple{\theta,R,h})$ \\
 \> $\theta^* \leftarrow \text{\tt optimize}(D, \theta^*)$ \\
 {\bf Output:} hypothetical optimal policy $\theta^*$
\end{tabbing}
\end{center}
\rule{\linewidth}{1pt}
\end{table}

Any policy search algorithm can now be combined with this proxy
environment to learn from scarce experience.  Table~\ref{table:learn}
shows a general reinforcement learning algorithm family using the proxy.
The definitions of {\tt pick\_sample}, {\tt add\_data}, and {\tt
optimize} are crucial to the behavior of the algorithm.  The {\reinf}
algorithm \cite{Williams92} is one particular instantiation of the {\tt
learn} routine where {\tt pick\_sample} returns $\theta^*$ without
consulting the data, {\tt add\_data} forgets all of the previous
data and replaces it with the most recent sample, and {\tt optimize}
performs one step of gradient descent (using $\hat V^{\rm IS}$ instead
of $\hat V^{\rm WIS}$).  The exploration extension to {\reinf} proposed
by \emcite{MeuleauTR00} is exactly the same
except the {\tt pick\_sample} routine now returns a policy that is a
mixture of $\theta^*$ and a random policy.

In order to make effective use of all of the data, we
define {\tt add\_data} to append the
new data sample to the collection of data.  This allows our algorithm to
remember all previous experience.  Additionally, we use an {\tt optimize}
routine that performs full optimization (not just a single step).
In {\reinf} and other policy search methods, the current policy guess
embodies all of the known information about the past (forgotten) samples.
It is therefore important to only take small steps of decreasing size to
insure the algorithm converges.  Because we now remember all of the
previous samples and we do not have any restraint on which policy we must
use for the next sample, we can search for the true optimum of the
estimator at every step.

\section{How to Sample?}

We still are left with a choice for the routine {\tt pick\_sample}.
This routine represents our balance between exploration and exploitation.
For this paper, we will consider a simple possibility to
illustrate this trade-off.  We let the {\tt pick\_sample} routine
have a single parameter $p^*$.  {\tt pick\_sample} is stochastic and with
probability $p^*$ it returns $\theta^*$.  The remainder of the time it
returns a random policy chosen uniformly over the space $\Theta$.  Thus,
the larger the value of $p^*$, the more exploitative the algorithm is.

\subsection{Illustration: Bandit Problems}
\label{example}

\begin{figure}[h]
\begin{center}
\begin{tabular}{c}
\psfig{file=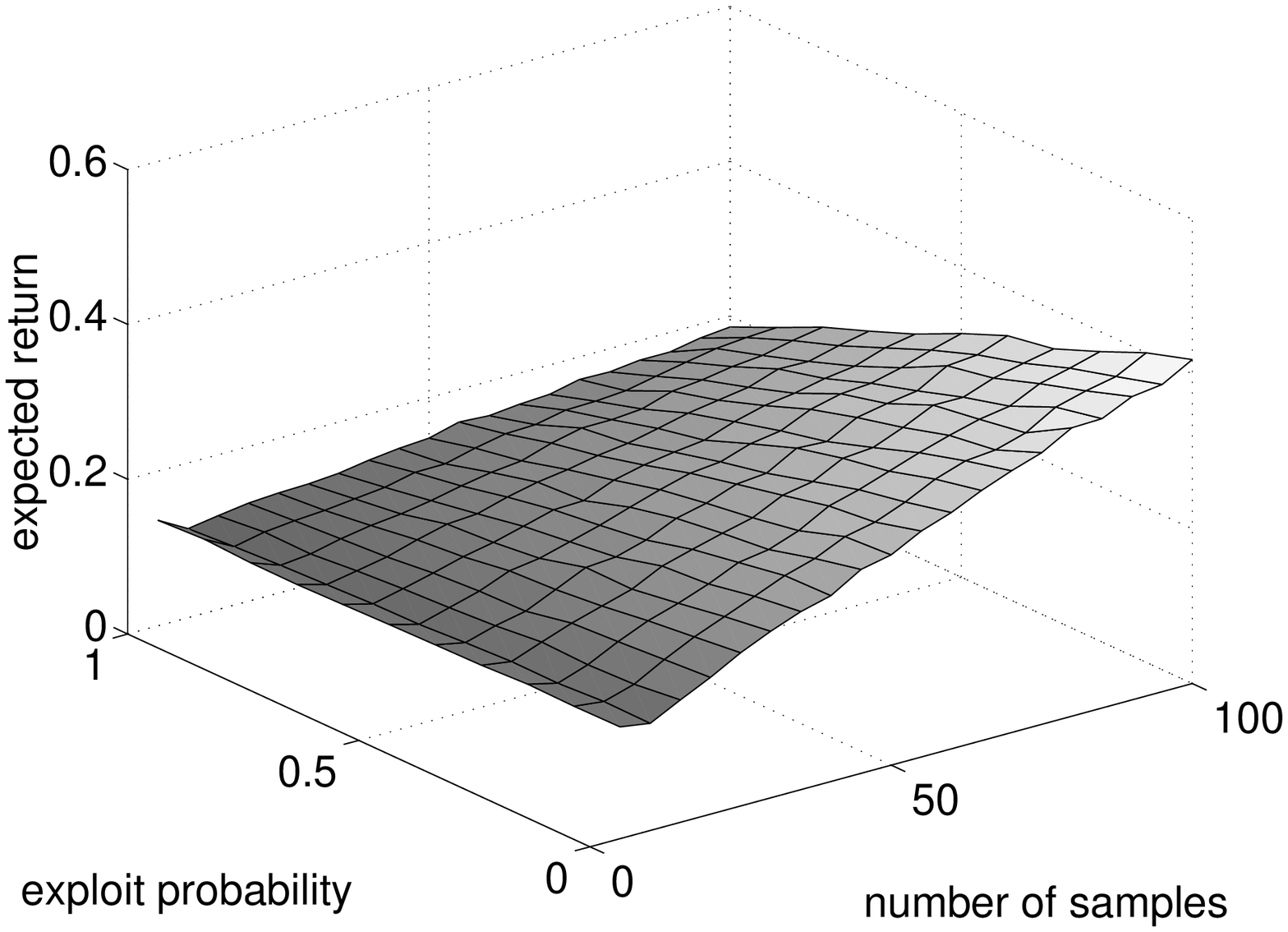,width=\figwidth,angle=0} \\
\psfig{file=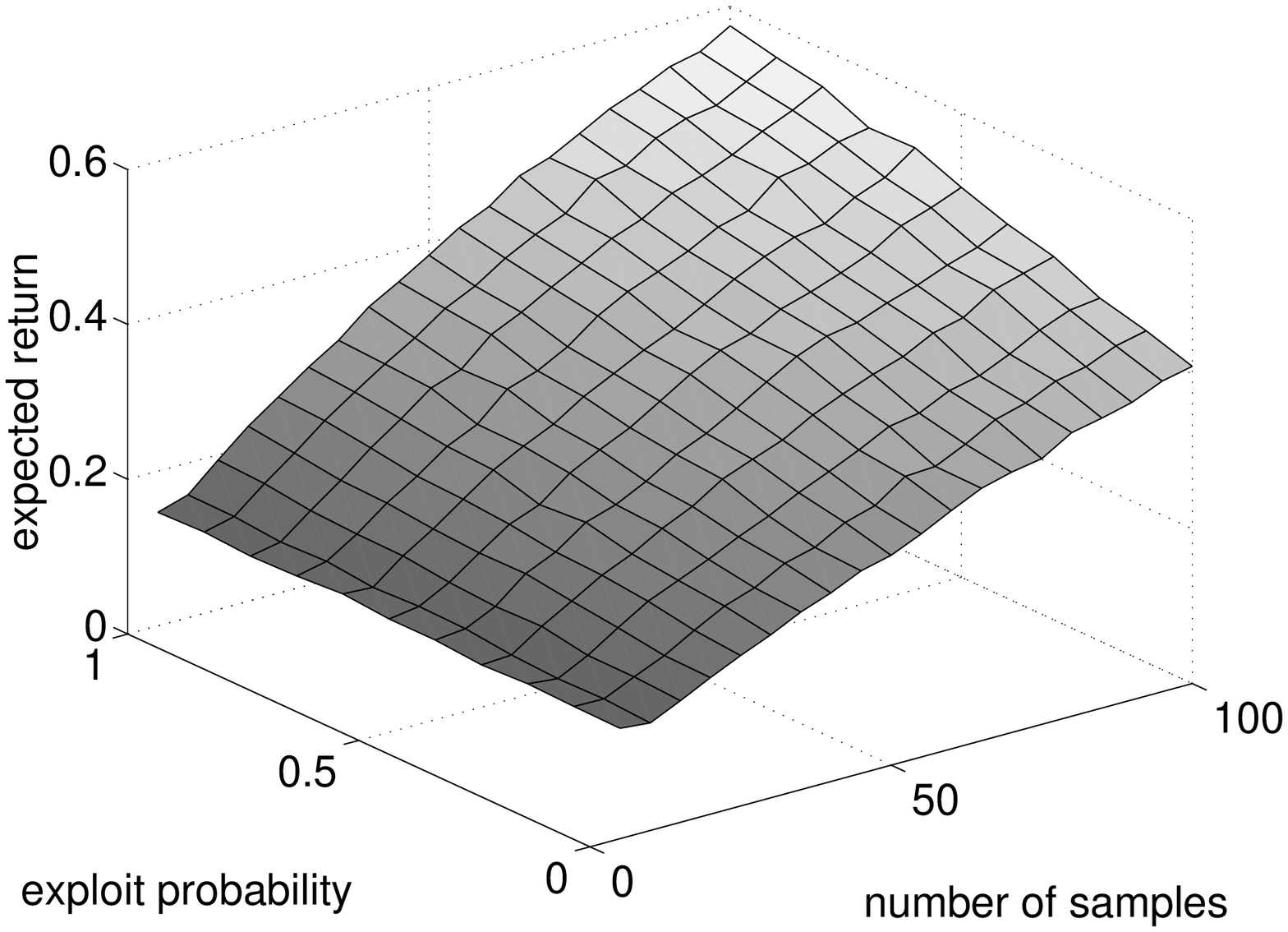,width=\figwidth,angle=0}
\end{tabular}
\end{center}
\caption{Results for the bandit problems. Top: ``HT'': hidden treasure.
Bottom: ``HF'': hidden failure.  Plotted is the expected return of the
resulting policy against the number of sample taken, $N$, and the
probability of exploiting, $p^*$.}
\label{bandit}
\end{figure}

Let us consider a trivial example of a bandit problem to illustrate
the importance of exploitation and exploration. The environment has a
degenerate state space of one state, in which two actions, $a_1$ and $a_2$,
are available. The space of policies available is stochastic and encoded
with one parameter, 
the probability of taking the first action, which
is constrained to be in the interval $[\underline{c},\overline{c}] =
[.1,.9]$.  We consider two problems, called ``HT'' (hidden treasure) and
``HF'' (hidden failure) both of which have the same expected returns for
actions: $1$ for $a_1$ and $0$ for $a_2$. In HF, $a_1$ always returns $1$,
while $a_2$ returns $10$ with probability $.99$ and $-990$ with probability
$.01$. In HT, $a_2$ always returns $0$, while $a_1$ returns $-10$ with
probability $.99$ and $+1090$ with probability $.01$. We would expect
a greedy learning algorithm to sample near policies that look better
under scarce information, tending to choose the sub-optimal $a_2$ in the
HT problem. This strategy is inferior to blind sampling, which samples
uniformly from the policy space and will discover the hidden treasure
of $a_1$ faster.
By contrast, for the HF problem we would expect
the greedy algorithm to do better by initially
concentrating on the $a_2$, which looks better, and discovering the hidden
failure sooner than blind sampling.

We ran the {\tt learn} algorithm from table~\ref{table:learn} for different
settings of the parameters $N$ and $p^*$.  Figure~\ref{bandit} shows the
true value of the resulting policies, averaged over $1000$ runs of the
algorithm.  While the plots may look discouraging, remember that these
problems are in some ways a worse-case situation.
The true value of the actions only becomes
apparent after sampling on the order of $100$ times.
The plots support our hypothesis about the relative success of 
exploitation.  However, although acting greedily is somewhat
better in HF, it is much worse in HT. This illustrates why, without any
prior knowledge of the domain and given a limited number of samples,
it is important not to guide sampling too much by optimization.

\section{How Much to Sample?}

\begin{table*}[t]
\caption{Comparison of sample complexity bounds.}
\label{bound_compare}
\begin{center}
\begin{tabular}{||l|l||}
\hline
Algorithm & Lower bound on sample complexity $N$ \\
\hline
likelihood ratio & $ \left(\dfrac{V_{\rm max}}{\epsilon}\right)^2 
      2^{4T}(1-\underline{c})^{4T} 
  \left( {\cal{K}}(\Theta) + \log(8/\delta) \right) $ \\
\hline
reusable trajectories & $\left(\dfrac{V_{\rm max}}{\epsilon}
           \right)^2 2^{2T} VC(\Theta) 
 \left( T + \log \left(\dfrac{V_{\rm max}}{\epsilon}\right) + \log(1/\delta)
 \right) \log(T) $ \\
\hline
\end{tabular}
\end{center}
\end{table*}

If we wish to guarantee with probability $1-\delta$ that the error in the
estimate of the value function is less than $\epsilon$, we can derive bounds
on the necessary sample size $N$, which depend on $\delta$, $\epsilon$,
$V_{max}$, and the
complexity of the hypothesis class expressed by the covering number ${\cal
N}$.  Our new result is an extension of the sample complexity
bound for the {\sc is} estimator~\cite{Peshkin01} to the {\sc wis}
estimator. We only quote the results here. The key point in the derivation is
the fact that
$$\sup_{\bar h, \bar h' \in H} \left| \hat V^{\rm WIS}_{\bar h}
- \hat V^{\rm WIS}_{\bar h'} \right| \leq \frac{V_{max}
\overline{c}^T}{N\underline{c}^T + \overline{c}^T}\,\,,
$$
where $\bar h'$ differs from $\bar h$ only
by one member trajectory $h_i$. Two inequalities
follow from this fact. Denote
$\eta = \max(\overline{c}^T,(1-\underline{c})^T)$.
The variance of the {\sc wis} estimator according to
Devroye's theorem is bounded as
$$
{\rm Var}\left\{ \hat V^{\rm WIS}_{\bar h}
\right\} \leq \frac{V^2_{max} \eta^4 N}{4(N + \eta^2)^2}
$$
and McDiarmid's theorem~\emcite{McDiarmid89} gives us a {\sc pac} bound (for
derivation see~\emcite{PeshkinPhD}):
\begin{multline*}
\delta = \Pr \left(\sup_{\theta \in \Theta} \left| \hat V^{\rm WIS}(\theta) - 
{\rm E} \left[\hat V^{\rm WIS}(\theta)\right]\right| > \epsilon \right) \\
\leq 4 {{\cal N}\left(\Theta,\frac{\epsilon}{8}\right)}
\exp \left[ -\frac{\epsilon^2 (N + \eta^2)^2}
{32 V^2_{max} \eta^4 N} \right]\;,
\end{multline*}
which gives a sample complexity bound very similar to these 
obtained by \emcite{KearnsTR99} (see next
section).
It is well known~\cite{Rubinstein81} that variance
of {\sc wis} estimate is $O(\frac{1}{N})$. The weak dependence on the
horizon $T$ is interesting and in accordance with empirical findings. The
covering number is defined through the value $V(\theta)$ and describes
the complexity of a policy class ({\it e.g.}~reactive policies or finite
state controllers) with respect to the structure of a reward function.

\subsection{Comparison to {\sc vc} Bound} 
\label{compVC}

The pioneering work by \emcite{KearnsNIPS99} considers the issue of
generating enough information to determine the near-best policy. We compare
our sample complexity results from above with a similar result for their
``reusable trajectories'' algorithm. Using a random policy (selecting actions
uniformly at random), reusable trajectories generates a set of history trees.
This information is used to define estimates that uniformly converge to the
true values. The algorithm relies on having a generative model of the
environment, which allows simulation of a reset of the environment to any
state and the execution of any action to sample an immediate reward. The
reuse of information is partial: the estimate of a policy value is built only
on the subset of experiences that are ``consistent'' with the estimated
policy.

We will make a comparison based on a sampling policy that selects one
of two actions uniformly at random:
$\Pr(a|h)=\frac{1}{2}$. For the horizon $T$, this gives us
an upper bound $\eta$ on the likelihood ratio:
$$
w_{\theta}(h,\theta') \leq 2^T(1-\underline{c})^T = \eta\,\,.
$$
Substituting this expression for $\eta$, we can compare our bounds to the
bound of \emcite{KearnsNIPS99} as presented in table~\ref{bound_compare}. The
metric entropy ${\cal {K}}(\Theta)$ takes the place of the {\sc VC} dimension
$VC(\Theta)$ in terms of policy class complexity. Metric entropy is a more
refined measure of capacity than VC dimension; the VC dimension is an upper
bound on the growth function which is an upper bound on the metric
entropy~\cite{Vapnik98}.

\subsection{Illustration: Load-Unload Problem}

\begin{figure}[ht]
\begin{center}
\begin{tabular}{c}
\begin{minipage}{2.4in}
\begin{center}
\setlength{\unitlength}{0.02in}
\begin{picture}(120,50)(0,-5)
\put(20,20){\circle{10}}
\matrixput(40,10)(20,0){4}(0,20){2}{\circle{10}}
\put(100,30){\circle*{10}}

\matrixput(56,13.77)(20,0){3}(0,20){2}{\vector(2,-1){0.46}}
\multiput(43.54,13.54)(20,0){3}{\qbezier(0,0)(1.46,1.46)(6.46,1.46)}
\multiput(50,15)(20,0){3}{\qbezier(0,0)(5,0)(6.46,-1.46)}
\multiput(43.54,33.54)(20,0){3}{\qbezier(0,0)(1.46,1.46)(6.46,1.46)}
\multiput(50,35)(20,0){3}{\qbezier(0,0)(5,0)(6.46,-1.46)}

\matrixput(44,6.23)(20,0){2}(0,20){2}{\vector(-2,1){0.46}}
\put(84,6.23){\vector(-2,1){0.46}}
\multiput(43.54,6.46)(20,0){3}{\qbezier(0,0)(1.46,-1.46)(6.46,-1.46)}
\multiput(50,5)(20,0){3}{\qbezier(0,0)(5,0)(6.46,1.46)}
\multiput(43.54,26.46)(20,0){2}{\qbezier(0,0)(1.46,-1.46)(6.46,-1.46)}
\multiput(50,25)(20,0){2}{\qbezier(0,0)(5,0)(6.46,1.46)}

\put(80,15.46){\vector(-1,-2){0.46}}
\qbezier(100,25)(100,20)(90,20)
\qbezier(90,20)(80,20)(80,15)

\put(34.54,30){\vector(4,1){.46}}
\qbezier(23.54,23.54)(30,30)(35,30)
\put(25.46,20){\vector(-4,-1){.46}}
\qbezier(36.54,26.54)(30,20)(25,20)

\put(24,16){\vector(-1,1){.46}}
\qbezier(35,10)(30,10)(23.54,16.46)

\put(14.54,20){\vector(4,1){.46}}
\qbezier(16.46,16.46)(15,15)(12.5,15)
\qbezier(12.5,15)(10,15)(10,17.5)
\qbezier(10,17.5)(10,20)(15,20)

\put(104,6){\vector(-2,1){.46}}
\qbezier(103.46,6.46)(105,5)(107.5,5)
\qbezier(107.5,5)(110,5)(110,7.5)
\qbezier(110,7.5)(110,10)(105,10)

\put(104,14){\vector(-1,-2){.46}}
\qbezier(103.54,13.54)(105,15)(105,20)
\qbezier(105,20)(105,25)(103.54,26.46)

\matrixput(32.5,0)(20,0){3}(0,40){2}{\line(1,0){15}}
\matrixput(32.5,0)(20,0){3}(15,0){2}{\line(0,1){40}}

\matrixput(7.5,0)(0,40){2}(85,0){2}{\line(1,0){20}}
\matrixput(7.5,0)(20,0){2}(85,0){2}{\line(0,1){40}}

\end{picture}
\end{center}
\end{minipage}
\end{tabular}
\end{center}
\caption{Diagram of the ``load-unload'' world. The agent observes its
position (box) but not whether the cart is loaded (node within the box). The
cart loads in the left-most state. If it reaches the right-most position
while loaded (upper path), it unloads and gets a unit of reward. The agent
has a choice of moving left or right at each position. Each trial begins in the
``load'' state and lasts $100$ steps. The optimal controller requires one bit
of memory.}
\label{fig:loadunload}
\end{figure}
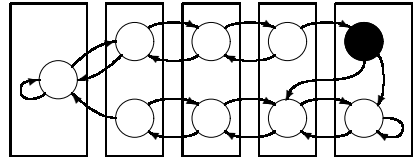

\begin{figure}[h]
\begin{center}
\psfig{file=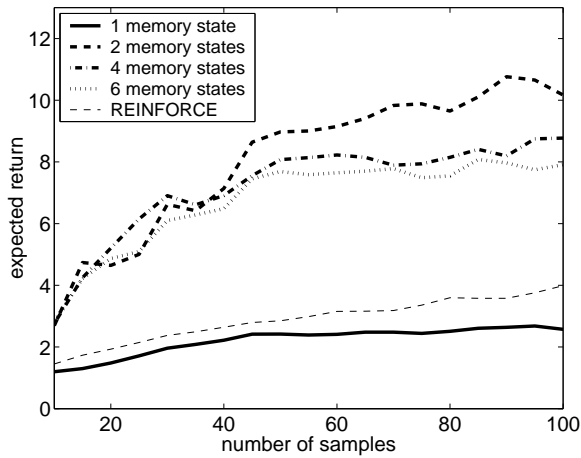,width=\figwidth,angle=0}
\end{center}
\caption{The expected return of the policy found by the algorithm as a
function of the number of samples for the environment shown in
figure~\protect\ref{fig:loadunload}. The exploitation parameter, $p^*$, was
set to $0.5$. These results are averaged over $80$ separate runs for each
number of samples. The solid line plots the performance using a reactive
policy (no memory). The dashed and dotted lines are for policies with
differing amounts of memory. Notice that while the reactive policy class
converges more quickly, its optimum is much lower. All of the other policy
classes have the potential to converge to the optimal return of $13$.
However, increased policy complexity results in slower learning.}
\label{mapg}
\end{figure}

The complexity of the problem is measured by the covering number,
${\cal N}$.  It encodes the complexity of the combination of the
{\pomdp} and the policy class.  We use the load-unload problem of
figure~\ref{fig:loadunload} to illustrate the effect of policy complexity.
The agent is a cart designed to shuttle loads back and forth between two
end-points on a line. The cart does not have sensors to indicate whether
it is loaded or unloaded, but it can determine its position on the line.
The optimal policy is one where the cart moves back and forth between
the leftmost and rightmost states moving as many loads as possible.
To do so requires some form of memory.  For our example, we will use
finite-state controllers with fixed memory sizes~\cite{SheltonPhD,PeshkinPhD}.

A finite-state controller is a class of policies with a fixed memory size.
The controller has its own internal memory state that is restricted to one
of a finite number of values.  At each time step, it not only selects which
action to take, but also a memory state for the next time step.  The
controller's choice of action and next memory state are independent
of the past given the current observation and memory state.  This model is
an extension of the reactive policy class to allow the controller to
remember a small amount about the past. Finite-state controllers have the
capability of remembering information for an arbitrarily long period of
time. 

Figure~\ref{mapg} demonstrates the effect of policy complexity on the
performance of the algorithm. This plot is the same as the ones in
figure~\ref{bandit} except that the exploitation probability, $p^*$, has been
fixed at $0.5$. The four lines depict results for different policy classes,
$\Theta$. The solid line is for reactive policies (policies with no memory)
whereas the dashed and dotted lines are for finite-state controllers with
varying amounts of memory. Only one bit of memory is required to perform
optimally in this environment. Using more than two states of memory is
superfluous. We can see that the simpler the policy class, the more quickly
the algorithm converges. However, with too simple a policy class ({\it i.e.}
reactive policies), the convergence is to a suboptimal policy. For
comparison, the thin dashed line presents the behaviour of {\sc reinforce}
with two internal state controller. As we have seen {\sc reinforce} forgets
past experience and picks up very slowly with the size of experience.

\section{Discussion}

Likelihood-ratio estimation seem to show promise in using data efficiently.
The {\tt pick\_sample} routine we present here is only one (simplistic) 
method of balancing exploitation and exploration.
More sophisticated methods including
maintaining a distribution over the space of policies might allow for 
a better balance and the possibility of learning a
useful sampling bias in a policy space for a particular application
domain and transferring it from one learning problem in that domain to
another. 
In general, estimating the variance of the proxy evaluator could aid in
selecting new samples for either exploration or exploitation.


Where {\reinf} keeps only the most recent sample, our algorithm keeps all
of the samples.  If a large amount of data is collected, it may be
necessary to employ a method between these two extremes and remember a
representative set of the samples.  Deciding which samples to 
``forget'' would be a difficult, but crucial, task.

Sample complexity bounds depend on the covering number as a measure of the
complexity of the policy space. Estimating the covering number is a
challenging problem in itself. However it would be more desirable to find a
constructive solution to a covering problem in a sense of universal
prediction theory ~\cite{Merhav98,Shtarkov87}. Obviously, given a covering
number there might be  several ways to cover the space. Finding a covering
set would be equivalent to reducing a global optimization problem to an
evaluation of several representative policies.

Another way to use sample complexity results is to find what is the minimal
experience necessary to be able to provide the estimate for any policy in the
class with a given confidence. This would be similar to the
structural risk minimization principal by Vapnik~\emcite{Vapnik98}. The
intuition is that given very limited data, one might prefer to search a
primitive class of hypotheses with high confidence, rather than to get lost
in a sophisticated class of hypotheses due to low confidence.

\subsubsection*{Acknowledgments}
The authors would like to thank Leslie Kaelbling for helpful discussions and
comments on the manuscript. C.S. was supported by grants from ONR
contracts Nos.~N00014-93-1-3085 \&
N00014-95-1-0600, and NSF contracts Nos.~IIS-9800032 \& DMS-9872936 while
at MIT and by ONR contract N00014-00-1-0637 under the
MURI program ``Decision Making under Uncertainty'' while at Stanford.
L.P. was supported by ARO grant \#DAA19-01-1-0601.


\bibliography{scarce}
\bibliographystyle{mlapa}

\end{document}